\documentclass[conference]{IEEEtran}
\usepackage{amsmath,epsfig, algorithm, algorithmic}

\IEEEoverridecommandlockouts

\usepackage{tikz}
\usepackage{textcomp}
\usepackage{hyperref}
\usepackage{lipsum}

\newcommand\copyrighttext{%
  \footnotesize \textcopyright 2024 IEEE. Personal use of this material is permitted.
  Permission from IEEE must be obtained for all other uses, in any current or future 
  media, including reprinting/republishing this material for advertising or promotional 
  purposes, creating new collective works, for resale or redistribution to servers or 
  lists, or reuse of any copyrighted component of this work in other works.}
\newcommand\copyrightnotice{%
\begin{tikzpicture}[remember picture,overlay]
\node[anchor=south,yshift=10pt] at (current page.south) {\fbox{\parbox{\dimexpr\textwidth-\fboxsep-\fboxrule\relax}{\copyrighttext}}};
\end{tikzpicture}}

\makeatletter
\newcommand{\linebreakand}{%
  \end{@IEEEauthorhalign}
  \hfill\mbox{}\par
  \mbox{}\hfill\begin{@IEEEauthorhalign}
}
\makeatother

\author{
  \IEEEauthorblockN{Zina-Sabrina Duma}
  \IEEEauthorblockA{\textit{LUT University}\\
    Finland \\
    Zina-Sabrina.Duma@lut.fi}
  \and
  \IEEEauthorblockN{Tuomas Sihvonen}
  \IEEEauthorblockA{\textit{LUT University}\\
    Finland}
  \and
  \IEEEauthorblockN{Jouni Susiluoto}
  \IEEEauthorblockA{\textit{Jet Propulsion Laboratory}\\
    \textit{California Institute of Technology} \\
    USA }
  \linebreakand 
  \IEEEauthorblockN{Otto Lamminpää}
  \IEEEauthorblockA{\textit{Jet Propulsion Laboratory}\\
    \textit{California Institute of Technology} \\
    USA }
  \and
  \IEEEauthorblockN{Heikki Haario}
  \IEEEauthorblockA{\textit{LUT University}\\
    Finland}
  \and
  \IEEEauthorblockN{Satu-Pia Reinikainen}
  \IEEEauthorblockA{\textit{LUT University}\\
    Finland}
}

\title{Kernel-based retrieval models for hyperspectral image data optimized with Kernel Flows}

\begin{document}

\maketitle


\copyrightnotice

\begin{abstract}
Kernel-based statistical methods are efficient, but their performance depends heavily on the selection of kernel parameters. In literature, the optimization studies on kernel-based chemometric methods is limited and often reduced to grid searching. Previously, the authors introduced Kernel Flows (KF) to learn kernel parameters for Kernel Partial Least-Squares (K-PLS) regression. KF is easy to implement and helps minimize overfitting. In cases of high collinearity between spectra and biogeophysical quantities in spectroscopy, simpler methods like Principal Component Regression (PCR) may be more suitable. In this study, we propose a new KF-type approach to optimize Kernel Principal Component Regression (K-PCR) and test it alongside KF-PLS. Both methods are benchmarked against non-linear regression techniques using two hyperspectral remote sensing datasets.
\end{abstract}


\section{Introduction}

Hyperspectral imagery is unlocking possibilities for environmental monitoring. Environmental retrieval models that utilize remotely sensed data are already being used to track  water quality \cite{goyens2022using}, agriculture \cite{weiss2020remote}, vegetation health \cite{chadwick2020integrating}, and in climate observations \cite{beamish2020recent}. 

Deep learning has been proven to be successful for retrieval model creation from hyperspectral data \cite{ozdemir2020deep}. However, it comes with limitations. If for instance a large number of parameters are optimized or the network is too deep, the models are prone to overfitting. Another common drawback is that the interpretability of deep learning models is limited, and a significant amount of data is often needed to train an unbiased model. Such large and complete datasets to train accurate deep learning models are difficult to produce. Airborne or spaceborne hyperspectral datasets with synchronized ground measurements are challenging and costly to put together \cite{zheng2023shift}. Data needs to be collected adequately over space and time, which also requires obtaining field measurements over the full time period.

The size of the datasets required by traditional chemometric models, such as the partial least-squares regression (PLS), is, however, smaller. This comes at the cost of reduced ability to model non-linear relationships between spectra and biogeophysical quantities. When the input and output are highly collinear, simpler approaches such as the Principal Component Regression (PCR) may be more appropriate \cite{guven2019examination}. Kernelized versions of chemometric methods are able to achieve improved performance in the case of non-linear dependency modeling, but learning the kernel function and its parameters is a non-trivial task. In literature, this is achieved through, for example, kernel target alignment, feature space matrix or genetic algorithm \cite{duma2023kf}.

In this paper, we propose using Kernel Flows (KF) \cite{owhadi2019kernel} to optimize chemometric retrieval models. The benefits of KF include: (i) reduced data overfitting through its cross-validation approach, (ii) ease of usage, as it converges to correct parameters regardless of initial values, and (iii) the ability to achieve global optimal values instead of local minima. We extend our previous work on optimizing kernel parameters for K-PLS \cite{duma2023kf} to Kernel PCR (K-PCR), illustrated with two hyperspectral data cases: a soil moisture soft sensor and a vegetation trait model. A newly developed cross-validation error loss function \cite{lamminpaa2024forward} was utilized in adapting the KF workflow to K-PCR and was also tested on KF-PLS.

\section{Related work}
\label{sec:previous}

To optimize kernel-based K-PLS models, the authors have implemented KF to learn kernel functions from parameterized families. Originally developed for Gaussian Process Regression (GPR), KF uses cross-validation and stochastic gradient descent (SGD) to minimize a loss function that increases norm similarity between random batch and sub-batch partitions. It can also utilize momentum methods, like Nesterov or Polyak's momentum, for faster convergence.
\begin{table}[H]
    \centering
    \caption{Kernel functions evaluated in the study. \label{tab:kerneltable}}
    \begin{tabular}{c|c}
        \textbf{Kernel} &  \textbf{Kernel Function}\\
        \hline
        Gaussian &  $exp \left( - \frac{|| \mathbf{x} - \mathbf{y} || ^ 2}{2 \sigma^2} \right)$ \\[0.2cm]
        Cauchy & $ \frac{1}{1 + \frac{|| \mathbf{x} - \mathbf{y} ||^2}{\sigma^2}}$\\[0.4cm]
        Matern1/2 & $ exp \left(- \frac{|| \mathbf{x} - \mathbf{y} ||}{\sigma} \right)$\\[0.2cm]
        Matern3/2 & $ \left(1 + \frac{\sqrt{3}|| \mathbf{x} - \mathbf{y} ||}{\sigma}\right) exp\left(- \frac{\sqrt{3}|| \mathbf{x} - \mathbf{y} ||}{\sigma}\right)$\\[0.2cm]
        Matern5/2 & $ \left(1 + \frac{\sqrt{5}|| \mathbf{x} - \mathbf{y} ||}{\sigma} + \frac{5|| \mathbf{x} - \mathbf{y} ||^2}{3\sigma^2}\right) \cdot$ \\[0.1cm] 
        & $exp\left(- \frac{\sqrt{5}|| \mathbf{x} - \mathbf{y} ||}{\sigma}\right)$\\
    \end{tabular}
    \label{tab:kernel}
\end{table}

In the list below, let S stand for startup, and I for iteration. The steps to optimize kernel parameters with KF are summarized as in \cite{duma2023kf}.
\begin{enumerate}
    \item[S1] Select a kernel function or a combination of kernel functions $k(x, x, \theta)$, as seen in Table \ref{tab:kerneltable} and initialize the kernel parameters $\theta_0$. Then, for each iteration $i = 1 .. n$,
    \item[I1] Draw a random batch 
    $\%_{batch}$ 
    of the data, $\mathbf{X}_{b}$, and map it using the selected kernel function $ K_{b} \gets k(\mathbf{X}_b, \mathbf{X}_b, \theta_i)$.
    \item[I2] Compute the weights for the batch $\mathbf{b}_{b} \gets \mathbf{K}_b \backslash \mathbf{Y}_b $ and the batch norm $n_{b} \gets \mathbf{b}_{b}^T \mathbf{K}_b  \mathbf{b}_{b}^T$.
    \item[I3] Draw a pre-selected number of subbatches  $\mathbf{X}_{s}$ from $\mathbf X_b$ and map it using the selected kernel function $ \textbf{K}_{s} \gets k(\mathbf{X}_s, \mathbf{X}_s, \theta_i)$. Then, for each sub-batch,
    \item[I4] Compute the weights for the batch $\mathbf{b}_{s} \gets \mathbf{K}_s \backslash \mathbf{Y}_s $ and the batch norm $n_{s} \gets \mathbf{b}_{s}^T \mathbf{K}_s  \mathbf{b}_{s}^T$. 
    \item[I5] Compute the loss function contribution $\rho_i \gets 1 - n_s / n_b$.  
    \item[I6] After iterating over sub-batches, average the iteration's loss values $\bar{\rho}$ and compute the gradient components with respect to each kernel parameter $\nabla(\theta_i, \bar{\rho})$.
    \item[I7] Update the parameters.
\end{enumerate}

Optimizing K-PLS performance with KF was demonstrated in \cite{duma2023kf}. In the KF-PLS approach, the weights computed with GPR are replaced with the regression coefficients computed with PLS.

\section{Proposed mathematical methods}
\label{sec:proposed}

In the present study, we are extending the KF approach for learning kernel parameters to various chemometric-based methods and propose an alternative loss function to enhance performance in this scenario.

In \cite{rosipal2001kernel}, the prediction equation in the case of K-PCR is described as
\begin{equation}\label{eq:trainPCR}
\begin{split}
    f(\textbf{X},\textbf{b}, \theta) & = \sum^{H}_{k=1} v_k \beta_k(x) + b_0 =\\ 
    & = \sum^{H}_{k=1} v_k \sum^{N}_{i=1} \lambda_k^{-1\slash2} u_i^{-k} \widetilde{k}(\textbf{x}_i,\textbf{X},\theta) + b_0  =\\
    & = \sum^{N}_{i=1} b_i \widetilde{k}(\textbf{x}_i,\textbf{X},\theta) + b_0 
\end{split}
\end{equation}
where $H$ is the number of non-linear principal components, $N$ is the number of samples present in model calibration, $\widetilde{k}()$ is the function that outputs a centered kernel matrix, $v_k$ is the regressor in latent space for the $k$-th component, $b_i$ is a regressor for the $i$-th sample, $b_0$ is the bias term, $\lambda$ represents the eigenvalues, and $u_i$ denotes the eigenvectors of the $i$-th sample. The centering is expressed as
\begin{equation}
    \widetilde{k}(\textbf{X}, \textbf{X}, \theta) = (\mathbf{I} - \frac{1}{N}\mathbf{1}_N\mathbf{1}_N^T) k(\textbf{X}, \textbf{X}, \theta) (\mathbf{I} - \frac{1}{N}\mathbf{1}_N\mathbf{1}_N^T)
\end{equation}
where $1_N$ is a vector of size $N$ with every value representing $1/N$.

For newly collected spectra $\mathbf{X}_{new} $, the model predictions can be computed with
\begin{equation}\label{eq:testPCR}
    f(\textbf{X}, \textbf{X}_{new},\textbf{b}, \theta) =  \sum^{N_t}_{i=1} b_i \widetilde{k}(\textbf{x}_i,\textbf{X}_{new},\theta) + b_0
\end{equation}
where $N_t$ is the number of new samples. The centering for the new kernel is done in a similar way
\begin{equation}
\begin{split}
    \widetilde{k}(\textbf{X}_{new}, \textbf{X}, \theta)  = & \left(k(\textbf{X}_{new}, \textbf{X}, \theta) - \frac{1}{N}\mathbf{1}_{N_t}\mathbf{1}_N^Tk(\textbf{X}, \textbf{X}, \theta)\right) \\  & \quad \times \left(\mathbf{I} - \frac{1}{N}\mathbf{1}_N\mathbf{1}_N^T\right).
\end{split}
\end{equation}
Here, \textbf{I} is the identity matrix, and $N_t$ is the number of new data points.
 
In PLS, the latent variables in the kernel space are rotated towards maximum covariance of \textbf{K} and \textbf{Y}. This step encodes dependencies between inputs and outputs, after which utilizing the original KF loss that enforces regularity of the regressor is a justified approach for learning the kernel parameters. In practice, this means that the batch and sub-batch norms should be similar (see step I5 above).

On the other hand, the principal components (PCs) in the case of K-PCR are not rotated towards maximum covariance. The first \textit{H} principal components in the batch can have little co-variance with the response. The explained variance in \textbf{Y} can be low both in the batch and the sub-batch, resulting in potentially similar norms (low loss value) even if the model would otherwise be incorrect. In training, this loss will force the difference to be small, which we have observed in practice lead to a faulty regression model.

To remedy this issue, we adopt a novel loss function for KF parameter optimization proposed by \cite{lamminpaa2024forward}. The method is adaptable to the optimization of KF-PCR, as described in Algorithm \ref{alg:KFPCR}, as it is based on prediction error minimization in a Leave-One-Out (LOO) cross-validation manner. This ensures learning a model that encodes correlations between inputs and responses. The final approach can be summarized in the following algorithm:
\begin{algorithm}[H]\label{alg:KFPCR}
    \caption{Kernel parameter optimization with modified Kernel Flows for K-PCR.}
    \textbf{\textit{Input:}} Initial kernel parameters ($\theta_0$), spectra matrix ($\mathbf{X}$), geobiophysical quantity ($\mathbf{y}$), number of PCs ($H$), number of iterations ($I$), number of samples\slash iteration ($N$), learning rate ($\alpha$) \textbf{\textit{Output:}} Optimized kernel parameters ($\theta_I$).
    \begin{algorithmic}[1]
    \FOR{$i = 1$ to $I$}
        \STATE $\mathbf{X}^{iter}, \mathbf{y}^{iter} \gets  \mathbf{X}_{\pi, \cdot}, \mathbf{y}_{\pi}$ where $\pi$ is an $n$ sample extraction from a random permutation.
        \FOR{$n = 1$ to $N$}
            \STATE $\mathbf{X}^{train}, \mathbf{y}^{train} \gets \mathbf{X}^{iter}_{\left[1:n-1, n+1:N\right], \cdot}, \mathbf{y}^{iter}_{\left[1:n-1, n+1:N\right]}$
            \STATE $ \mathbf{K}^{train} \gets k(\mathbf{X}^{train},\mathbf{X}^{train})$
            \STATE $\widetilde{\mathbf{K}}^{train} \gets (\mathbf{I} - \frac{1}{N}\mathbf{1}_N\mathbf{1}_N^T)\mathbf{K}^{train} (\mathbf{I} - \frac{1}{N}\mathbf{1}_N\mathbf{1}_N^T)$
            \STATE $ \mathbf{T} \gets kpca(\widetilde{\mathbf{K}}^{train}, H) $
            \STATE $ \mathbf{b} \gets \mathbf{T} \backslash \mathbf{y}^{train}$
            \STATE $ \rho_n \gets ||\mathbf{y}^{iter}_n - f(\mathbf{X}^{train}, \mathbf{X}^{iter}_n, \mathbf{b},\theta_i)||_2$ See \textit{f} in Eq. \ref{eq:testPCR}.
        \ENDFOR
        \STATE $\bar{\rho}_i \gets \sum_{n=1}^N \rho_n$
        \STATE $\nabla_{\theta, i} \gets diff(\theta_i, \bar{\rho}_i)$
        \STATE $\theta_{i-1} \gets \alpha \nabla_{\pmb \theta, i} f(\theta_i)$
    \ENDFOR
    \end{algorithmic}
\end{algorithm}

Instead of the parameter update in Alg. 1, one can utilize for faster convergence an updating momentum, such as Polyak's or Nesterov's momentum. 

\section{Case studies}
\label{sec:case studies}

Two hyperspectral imaging case studies have been chosen to illustrate the method. 

\begin{figure}
\begin{minipage}[b]{1\linewidth}
  \centering \centerline{\epsfig{figure=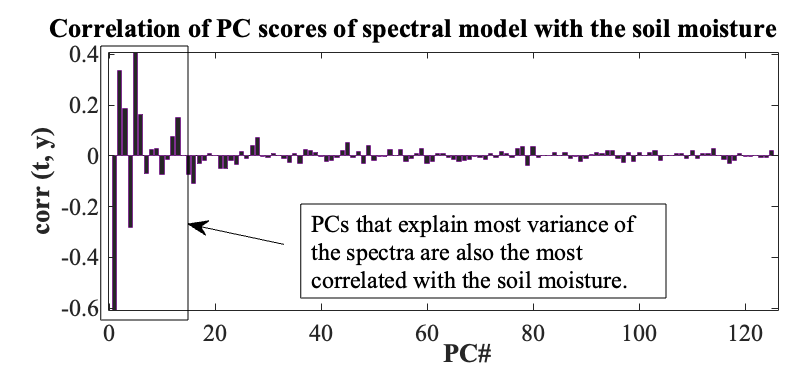,width=\textwidth}}
\end{minipage}
\caption{PCR is preferred to PLS when spectra are highly correlated with the response. The soil moisture dataset has a high correlation to spectral profiles.}
\label{fig:correlation}
\end{figure}

\textbf{Case I}. Hyperspectral soil imagery to calibrate moisture soft-sensors. The dataset presented by Riese \& Keller, 2018 \cite{riesekeller2018} contains 450 - 950 \textit{nm} spectra with a spectral resolution of 4 \textit{nm}. The modelling goal of \textit{Case I} is to demonstrate the KF optimization and retrieval model performance with KF-PCR. The PCR is preferably utilized in scenarios in which (a) the variables have a high correlation to the response variable, not justifying the utilization of more complex methods, and (b) the data is high-dimensional. Figure \ref{fig:correlation} showcases the correlation between PCs in the original space and the response variable. 

\begin{figure}[htb]
\begin{minipage}[b]{.48\linewidth}
  \centering \centerline{\epsfig{figure=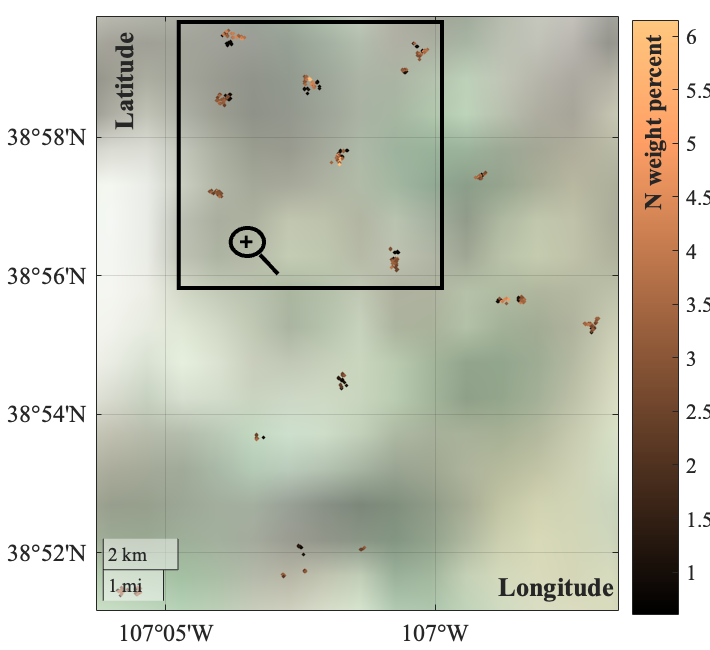,width=5.0cm}}
  \centerline{(a)}\medskip
\end{minipage}
\hfill
\begin{minipage}[b]{0.48\linewidth}
  \centering
\centerline{\epsfig{figure=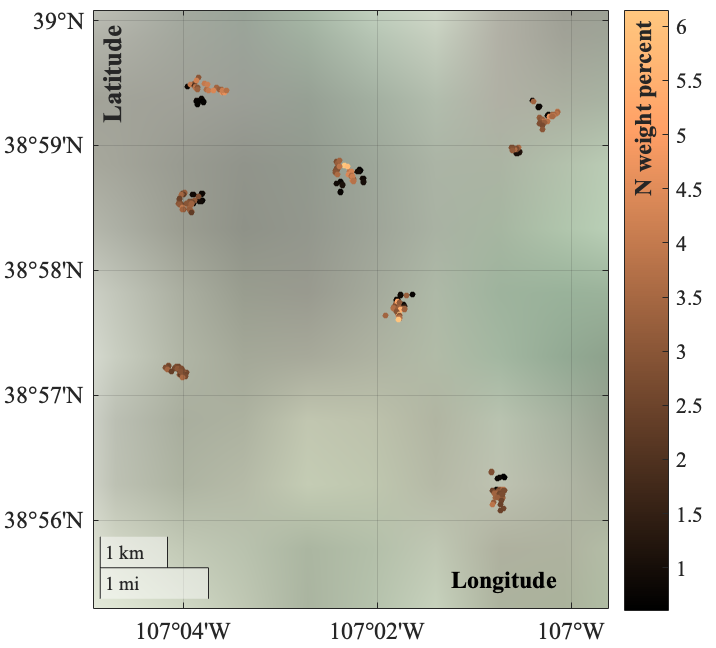,width=5.0cm}}
  \centerline{(b)}\medskip
\end{minipage}
\caption{Plant trait values and distribution for the Nitrogen weight content [\%] \cite{chadwick2020integrating}.}
\label{fig:data}
\end{figure}

\textbf{Case II}. Hyperspectral vegetation airborne imagery to calibrate plant trait soft sensors. The dataset introduced in Chadwik \textit{et al.}, 2020 \cite{chadwick2020integrating} was originally modelled with PLS. The goal of the case is to evaluate if KF-PLS with the proposed loss function is competitive compared to other non-linear methods. The trait presented in the results is illustrated in Fig. \ref{fig:data}. The mass percentage of Nitrogen in the needle trees has been chosen for showcasing due to its low performance in the linear methods \cite{chadwick2020integrating}.
\section{Results and Discussion}
\label{sec:results}
\begin{figure}
\begin{minipage}[b]{1\linewidth}
  \centering \centerline{\epsfig{figure=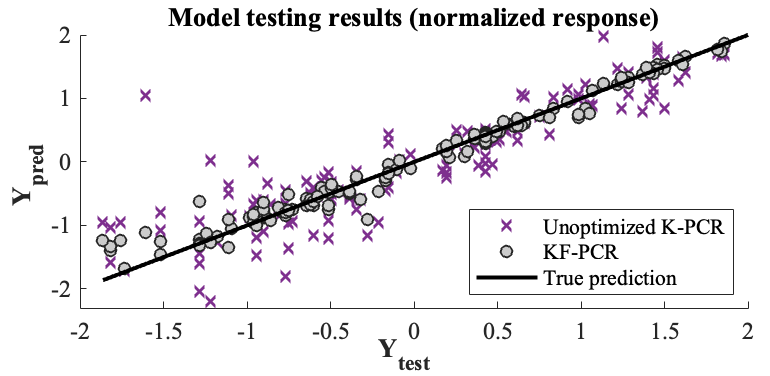,width=9.0cm}}
\end{minipage}
\caption{The test partition results after K-PCR optimization with Kernel Flows, when compared to the unoptimized K-PCR with the kernel parameters $\theta = 1$.}
\label{fig:response}
\end{figure}

\textbf{Case I.} 
The first case study showcases a scenario in which the response variable (soil moisture) is highly correlated with the response variable. As seen in Figure \ref{fig:response}, the unoptimized K-PCR model has higher residuals for test partition prediction. In the unoptimized version, all kernel parameters have the value '1'. A combination of kernels is able to minimize the residuals and achieve better precision with only one nonlinear PC.

\begin{figure}
\begin{minipage}[b]{1\linewidth}
  \centering \centerline{\epsfig{figure=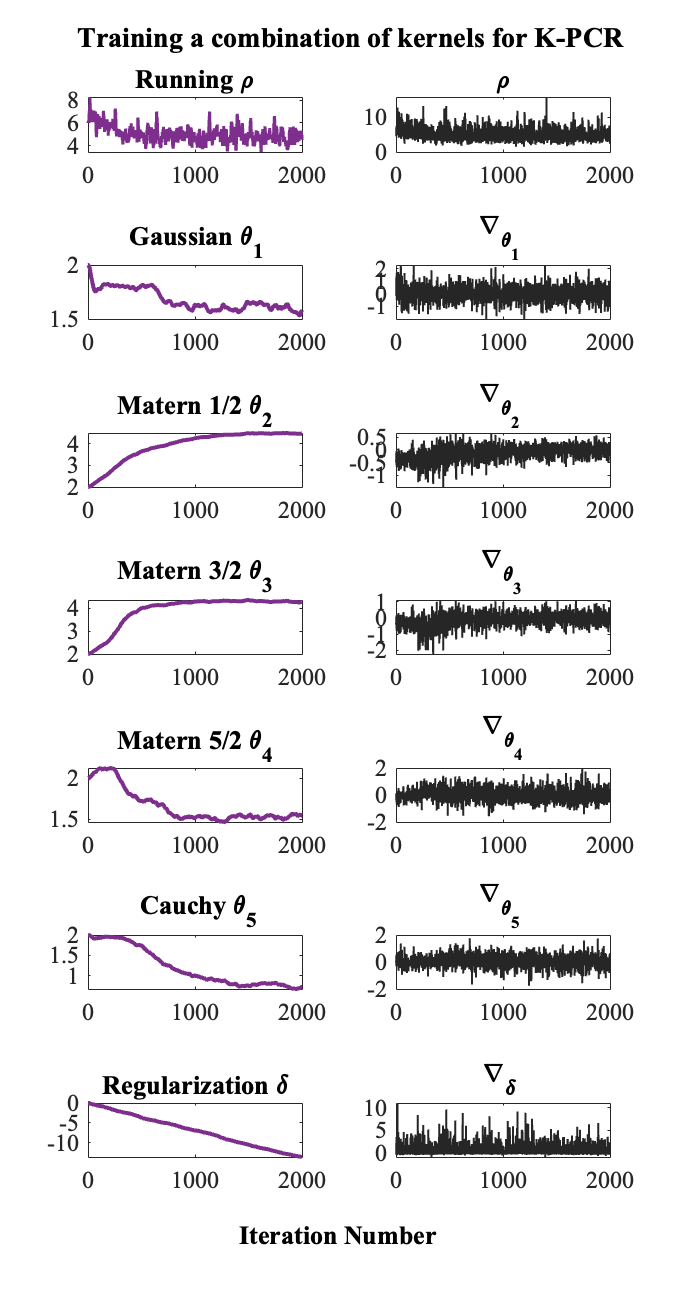,width=8.0cm}}
\end{minipage}
\caption{Parameter learning for a combination of kernels. One kernel parameter (width) was learned for each kernel in Table \ref{tab:kernel}. To keep the values in the positive range, the exponential of the parameters is utilized in learning.}
\label{fig:training}
\end{figure}

In Figure \ref{fig:training}, the learning process for a combination of kernels is presented. Compared to the training process using the traditional KF loss, \cite{duma2023kf}, the training process is less stable, with a larger variation in the loss. Some parameters converge into a steady value, while others have a slight variation around the convergence value. This occurs due to (i) the kernel parameters being correlated to each other or (ii) the kernel not having an important role in the combination of kernels. The mean of the gradient component for the kernel parameters approaching convergence is 0 for all converged parameters. 

\textbf{Case II.}
KF-PLS emerged as the top performer, followed by GPR. The GPR model, utilizing the Rational Quadratic kernel, had a parameter for each wavelength and took over 3000 seconds to train, while KF-PLS only fitted only two parameters and trained in 107 seconds. KF-PLS and KF-PCR outperformed linear methods due to their ability to capture non-linearity in the dataset. Additionally, they reduced the risk of overfitting, outperforming some of the non-linear methods, by learning fewer parameters and due to the the cross-validative nature of KF. When compared to the original KF loss \cite{owhadi2019kernel}, the updated KF loss brought a +33\% performance improvement in the KF-PCR, and a +2\% improvement in the KF-PLS results.

\begin{table}
\centering
\caption{The model performance of the optimized evaluated methods compared to the new loss applied to KF-PLS and proposed K-PCR method. All the compared methods have been optimized as well in MATLAB's \textit{Regression Learner}.}
\label{tab:resultsComparison}
\begin{tabular}{c|c|c}
\textbf{Method }                & \textbf{Optimized }         & $R^2_{test}$ \\
\hline
\textbf{KF-PLS} (modified $\rho$)      & \textbf{Matern5/2 }         &\textbf{ 0.580}                                 \\
KF-PCR                 & Cauchy             & 0.481                                 \\
PLS (Original article) & Ref.~\cite{chadwick2020integrating}              & 0.201                                 \\
Linear Regression      & -                  & 0.331                                 \\
Tree                   & Coarse             & 0.290                                 \\
SVM                    & Cubic              & 0.518                                 \\
Efficient Linear       & Least-Squares      & 0.023                                 \\
Ensemble               & Bagged Trees       & 0.459                                 \\
GPR                    & Rational Quadratic & 0.531                                 \\
Neural Network         & Bilayered          & 0.359                                
\end{tabular}
\end{table}

\section{Conclusions}
\label{sec:print}

In this paper, we propose a method to optimize the kernel parameters for K-PCR using a modified KF approach. Our results demonstrate an improvement in soil moisture prediction from hyperspectral airborne data by utilizing a combination of kernels. We also tested this new approach with KF-PLS in a case study focused on estimating nitrogen weight content in plants from airborne spectra, where the spectra and responses were not highly collinear. The results consistently outperformed other linear and non-linear methods, indicating that our approach is effective for optimizing KF-PLS as well.

The method shows great potential for various hyperspectral data analysis. Apart from the illustrated environmental retrieval models on soil moisture and plant traits, the models can be extended to water quality retrieval models. Biogeophysical quantity maps can be generated from the models with more accuracy, as the proposed kernel-based models do not overfit the training data. The optimized kernel methods also have the potential of replacing traditional PLS in hyperspectral pansharpening models. 

This study has limitations related to the selection of training data for model calibration. Larger training datasets lead to increased kernel matrices and longer training times. Future work could explore sample selection strategies for optimal kernel construction following parameter optimization.
\section{Acknowledgements}
Funding from Research Council of Finland for Centre of Excellence of Inverse Modelling and Imaging, project number 353095, is acknowledged. The Research Council of Finland, through the Flagship of Advanced Mathematics for Sensing, Imaging and Modelling, decision number 359183, is also acknowledged. The research effort by JS and OL was carried out at the Jet Propulsion Laboratory, California Institute of Technology, under a contract with the National Aeronautics and Space Administration (80NM0018D0004).
\bibliographystyle{IEEEbib}
\bibliography{strings,refs}

\end{document}